\pdfoutput=1
\documentclass[11pt]{article}

\usepackage{acl}

\usepackage{times}
\usepackage{latexsym}
\usepackage{amsfonts}

\usepackage[T1]{fontenc}
\usepackage[utf8]{inputenc}

\usepackage{graphicx}
\usepackage{subcaption}

\usepackage{microtype}
\usepackage{cinzel} %inserted by steven to add MERLOT font

\usepackage{multirow}
\usepackage{tabularx}

\usepackage{comment}
\usepackage{wrapfig} 
\usepackage{calligra}

\usepackage{aurical}

\newcommand{\twoline}[2]{\begin{tabular}[t]{@{}c@{}}#1\\#2\end{tabular}}

\makeatletter
\def\thickhline{%
  \noalign{\ifnum0=`}\fi\hrule \@height \thickarrayrulewidth \futurelet
   \reserved@a\@xthickhline}
\def\@xthickhline{\ifx\reserved@a\thickhline
               \vskip\doublerulesep
               \vskip-\thickarrayrulewidth
             \fi
      \ifnum0=`{\fi}}
\makeatother

\makeatletter
\newcommand\footnoteref[1]{\protected@xdef\@thefnmark{\ref{#1}}\@footnotemark}
\makeatother

\newlength{\thickarrayrulewidth}
\usepackage{amssymb}% http://ctan.org/pkg/amssymb
\usepackage{pifont}% http://ctan.org/pkg/pifont

\usepackage{tipa}
\usepackage{makecell}
\usepackage{enumitem,kantlipsum}

\title{Exploring Euphemism Detection in Few-Shot and Zero-Shot Settings}

\author{Sedrick Scott Keh \\ 
Carnegie Mellon University \\
\texttt{skeh@cs.cmu.edu}}

\begin{document}
\maketitle

\begin{abstract}
This work builds upon the Euphemism Detection Shared Task proposed in the EMNLP 2022 FigLang Workshop, and extends it to few-shot and zero-shot settings. We demonstrate a few-shot and zero-shot formulation using the dataset from the shared task, and we conduct experiments in these settings using RoBERTa and GPT-3. Our results show that language models are able to classify euphemistic terms relatively well even on new terms unseen during training, indicating that it is able to capture higher-level concepts related to euphemisms.
\end{abstract}

\section{Introduction}
\label{sec:intro}
Euphemisms are figures of speech which aim to soften the blow of certain words which may be too direct or too harsh \cite{magu-luo-2018-determining, felt-riloff-2020-recognizing}. In the EMNLP 2022 FigLang Workshop Euphemism Shared Task, participating teams are given a set of sentences with potentially euphemistic terms (PETs) enclosed in brackets, and the task is to classify whether or not the PET in a given sentence is used euphemistically. 

In this task/dataset, however, there are many PETs which are repeated throughout both the training and testing sets (more details in Section \ref{sec:dataset}). In addition, several PETs are classified as euphemistic almost 100\% of the time during training. This raises an important question: is the model actually learning to classify what a euphemism is, or is it simply reflecting back things it has seen repeatedly during training? How do we know if the model we train can truly capture the essence of what a euphemism is? Even among humans, this is a very nontrivial task. If one hears the phrase ``lose one's lunch'' for the first time, for example, it may not be immediately obvious that it is a euphemism for throwing up. However, when used in a sentence, the context clues together with an understanding of the meanings of the words ``lose'' and ``lunch'' will allow a human to piece together the meaning. For a machine to be able to do this, however, is not trivial at all.

To this end, we test this by checking whether a model can correctly classify PETs it has never seen during training. This leads us to our few-shot/zero-shot setting. The two key contributions of our paper are as follows: 1) We propose and formulate the few-shot and zero-shot euphemism detection settings; and 2) We run initial baselines on these euphemisms using RoBERTa and GPT-3, and we present a thorough analysis of our results.  

\section{Related Work}
\label{sec:related_work}
Compared to other figures of speech like similes \cite{chakrabarty-etal-2020-generating} and metaphors \cite{chakrabarty-etal-2021-mermaid}, work on euphemisms has been limited. 
% In the early days, studies focused on aspects such as politeness \cite{aubakirova-bansal-2016-interpreting} and formality \cite{pavlick-tetreault-2016-empirical} rather than directly working on euphemisms. 
Recently, \citet{cats-pets, lee-etal-2022-searching} released a new dataset of diverse euphemisms and conducted analysis on automatically identifying potentially euphemistic terms. In the past, \citet{felt-riloff-2020-recognizing} used sentiment analysis techniques to recognize euphemistic and dysphemistic phrases. Other studies also focused on specific euphemistic categories such as hate speech \cite{magu-luo-2018-determining} and drugs \cite{zhu2021selfsupervised}.

In terms of zero-shot figurative language detection, the existing literature has also been quite limited. The few existing studies \cite{Schneider22:MDL} mostly focus on metaphors and on low-resource settings. This leaves out less common figures of speech such as euphemisms, and the low-resource formulation is also not exactly identical to the zero-shot setting we explore in this paper.

\section{Task and Dataset}
\label{sec:dataset}
Our task is similar to the FigLang 2022 Workshop Shared Task on Euphemism Detection. Given a sentence containing a potentially euphemistic term (PET), we want to determine whether the PET is used euphemistically. The key difference with our task is that we perform the binary classification on a few-shot/zero-shot setting. Similarly, we use the dataset proposed by \citet{cats-pets}, which contains 1965 sentences with PETs, split across 129 unique PETs and 7 different euphemistic categories (e.g. death, employment, etc.) Furthermore, the dataset also contains additional information such as the category and the status of the PET (``always euph'' vs ``sometimes euph'').

\section{Methodology}
\label{sec:methodology}
\subsection{Constructing the Few-Shot Setting}
\label{subsec:few-shot-construction}
For the $k$-shot setting, we want the PETs in the validation/test set to have appeared in the training set only $k$ times. Let our set of PETs be $P = \{p_1, p_2, \dots p_N\}$. We construct the test set as follows. First, we randomly sample a PET $p_i$ from $P$, then find all sentences $s_1, s_2, \dots s_M$ containing PET $p_i$. Out of these $M$ sentences, we sample $k$ sentences $s_{j_1}, s_{j_2}, \dots s_{j_k}$ to keep in our training set, moving all the $(M-k)$ remaining sentences $s_j$ to our test set. We repeat this process until we reach the desired size for our validation/test set. In our case, we stop when the validation and test each reach around ~15\% of our entire dataset ($\pm2\%$ to account for the fact that it's unlikely to reach 15\% exactly). In practice, we sample 30\% for the validation+test set combined, then randomly split this 30\% into two sets of 15\% in order to increase the PET diversity in both the validation and the test splits. For the $k$-shot setting, we use $k=1$ and $k=3$. The dataset statistics for the $k$-shot datasets can be found in Table \ref{tab:dataset-statistics}.

\begin{table}[]
    \small
    \centering
    \begin{tabular}{c|cc}
         & \textbf{Ave. Test Size} & \twoline{\textbf{Ave. \# of unique}}{\textbf{PETs in test}} \\
         \hline
        Standard & 295.0 & 93.3 \\
         \hline
        Few-Shot (k=1) & 279.6 & 35.0 \\
        Few-Shot (k=3) & 281.2 & 35.4 \\
        \hline
        0-shot (random) & 280.6 & 34.3 \\
        \hline
        Death & 174.0 & 14.9 \\
        Sexual Activity & 45.0 & 10.4 \\
        Employment & 176.0 & 23.5 \\
        Politics & 161.0 & 20.9 \\
        Bodily Functions & 26.0 & 7.0 \\
        Physical/Mental & 299.0 & 36.0 \\
        Substances & 88.0 & 9.1 \\
    \end{tabular}
    \caption{Dataset statistics for the few-shot and zero-shot settings. Because there is some stochasticity involved in dataset creation, we take averages over 10 samples.}
    \label{tab:dataset-statistics}
\end{table}

\subsection{Constructing the Zero-Shot Setting}
\label{subsec:zero-shot-construction}
For the zero-shot setting, we want the PETs in the validation/test set to never have appeared in the training set. There are two ways to achieve this:

\begin{enumerate}[wide, labelwidth=!, labelindent=0pt]
\item \textbf{Random Sampling} -- The construction for this is similar to that of the few-shot setting, except here, we don't sample $s_{j_1}, s_{j_2}, \dots s_{j_k}$ to keep in the training set but rather move all $M$ sentences $s_1, s_2, \dots s_M$ to our validation/test set. 
\item \textbf{Type-based} -- Rather than randomly choosing assorted PETs to holdout into our test set, we instead choose the test set PETs to all come from a single category, while the training set will come from the remaining categories. These categories are provided alongside the sentences in the dataset by \citet{cats-pets}, and there are 7 categories in total. Because some categories may contain more sentences (and more PETs) than others, then the sizes of the training splits of these categories will be different. To address this, we subsample from the training splits of the categories with excess rows to match the training category with the least number of rows. This way, we ensure that all categories have an equal number of rows of training data, and so any changes in performance will be likely due to the data quality (rather than due to simply having more/less data). At the end, this gives us a training size of 1367 rows for each category. For the test splits, different categories also have different sizes, but we choose to leave the test split sizes unchanged and opted not to do the sampling like we did for the training step because the smallest testing category has size 26 (``bodily functions''), while some other categories had test sizes of 200+ (``physical/mental''), so we found it impractical to force the test sizes to be identical. Statistics for these datasets can be found in Table \ref{tab:dataset-statistics}. In theory, having larger test sets will mostly affect the variance, but the mean should not be affected that much. We comment more on this in Section \ref{sec:results_and_analysis}.
\end{enumerate}

\subsection{Models}
\label{subsec:base-models}
We consider two different types of baseline models. First, we consider networks which we can reasonably fine-tune. For this group, we select RoBERTa \cite{liu2019roberta}, covering both the RoBERTa-base model and the RoBERTa-large model, which have been extensively used for classification. The rationale behind choosing RoBERTa was twofold. First, RoBERTa is a commonly used standard for various classification tasks and has generally been shown to perform better than other simple transformer-based models such as BERT \cite{devlin-etal-2019-bert}. Second, it has empirically been shown to work sufficiently well when dealing with euphemisms, as \citet{lee-etal-2022-searching} used RoBERTa-based sentiment and offensiveness models to search for euphemisms.

In addition, we also try out large language models such as GPT-3 (davinci) \cite{gpt3}, which has been known to work well on zero-shot and few-shot settings. We are interested to find out whether the large-scale pretraining provides GPT-3 with the capability to implicitly model the concept of ``euphemism-hood'', which is built from several other adjacent concepts such as politeness and tone. We hence explore using both zero-shot and few-shot prompts for GPT-3.

\section{Experiment Setup}
\label{sec:experiment_setup}
\subsection{RoBERTa Implementation Settings}
\label{subsec:roberta-implementation}
For both RoBERTa-base and RoBERTa-large, we fine-tune for 10 epochs, taking the model with the best validation performance (F1) as our final model. For RoBERTa-base, we use a learning rate of 1e-5 and a batch size of 16, while for RoBERTa-large, we use a learning rate of 5e-6 and a batch size of 4. All other hyperparameters such as learning rate decay and warmup steps are according to the default settings of HuggingFace's trainer function.

\begin{table*}[t]
    \small
    \centering
    \begin{tabular}{c|c|ccc|ccc|ccc|}
        & & \multicolumn{3}{c}{\textbf{RoBERTa-base}} & \multicolumn{3}{c}{\textbf{RoBERTa-large}} & \multicolumn{3}{c}{\textbf{GPT-3 (davinci)}} \\
        & & P & R & F1 & P & R & F1 & P & R & F1 \\
        \hline
        Standard Model & - & 0.850 & 0.799 & 0.824 & 0.877 & 0.812 & 0.836 & - & - & - \\
        \hline
        \multirow{2}{*}{Few-Shot} & k=1 & 0.802 & 0.744 & 0.759 & 0.818 & 0.748 & 0.769 & 0.565 & 0.551 & 0.546 \\
        & k=3 & 0.834 & 0.795 & 0.808 & 0.879 & 0.798 & 0.825 & 0.624 & 0.599 & 0.617 \\
        \hline
        Zero-Shot (Random) & - & 0.770 & 0.699 & 0.715 & 0.798 & 0.726 & 0.740 & 0.537 & 0.543 & 0.507 \\
        \hline
        \multirow{7}{*}{Zero-Shot (Type-based)} & Death & 0.782 & 0.735 & 0.742 & 0.803 & 0.748 & 0.761 & 0.453 & 0.457 & 0.448 \\
        & Sexual Activity & 0.647 & 0.606 & 0.622 & 0.633 & 0.603 & 0.615 & 0.533 & 0.550 & 0.477 \\
        & Employment & 0.778 & 0.790 & 0.781 & 0.782 & 0.817 & 0.792 & 0.537 & 0.532 & 0.479 \\
        & Politics & 0.754 & 0.622 & 0.645 & 0.826 & 0.645 & 0.688 & 0.537 & 0.558 & 0.484 \\
        & Bodily Functions & 0.500 & 0.240 & 0.324 & 0.500 & 0.416 & 0.480 & 0.500 & 0.192 & 0.278 \\
        & Physical/Mental & 0.757 & 0.663 & 0.689 & 0.750 & 0.680 & 0.693 & 0.517 & 0.510 & 0.489 \\
        & Substances & 0.897 & 0.858 & 0.878 & 0.913 & 0.883 & 0.895 & 0.553 & 0.551 & 0.486 \\
    \end{tabular}
    \caption{Experiment results for RoBERTa-base, RoBERTa-large, and GPT-3 (davinci). Results are averaged over 5 experiments with different dataset splits.}
    \label{tab:main-results}
\end{table*}

\subsection{GPT-3 Implementation Settings}
\label{subsec:gpt3-implementation}
We use the largest version of GPT-3 (davinci). For the zero-shot settings, we prompt it with the phrase ``Is the word [PET] used euphemistically in the following sentence: [SENT]'', where [PET] and [SENT] represent the euphemistic term and current sentence in question. Here, we conduct a small amount of prompt engineering. For instance, we also tried out ``Does this sentence contain a euphemism: [SENT]'' or adding ``(Yes/No)'' before or after our current formulation. We found that our current formulation performs the best among these variations, which is why we choose to report that in Table \ref{tab:main-results}. Meanwhile, for few-shot settings, we simply repeated our zero-shot prompt, followed by either ``Yes'' or ``No'' corresponding to the label, and a line break to separate different examples.

Another key challenge with GPT-3 is mapping the responses to 0/1 binary classes. Because GPT-3 is a generative model, it may not necessarily just answer yes/no; instead, it may generate long paragraphs or unrelated characters. To do this mapping, we use a rule-based method. First, if the first 3 characters of the response is ``yes'' or if the first 2 characters are ``no'', then we can immediately map them. Next, we gather a list of ``1-class'' phrases and a list of ``0-class'' phrases. Here, ``1-class'' phrases include ``is a euphemism'', ``is used euphemistically'', ``can be considered a euphemism'', ``it seems like it'', etc. In other words, when these phrases appear in a sentence, then the label is most likely a 1. This likewise holds for ``0-class'' phrases which are indicative of the label being 0. This includes phrases such as ``not a euphemism'' or ``does not appear to be euphemistic''. Lastly, GPT-3 sometimes generates random noise, irrelevant sentences, or says something like ``I'm not sure'' or ``I can't answer that''. For these remaining cases, we choose to just ignore them from our scoring. Based on our experiments, this happened only around 4\% of the time, so we believe the change to be not that significant. The full list of ``1-class'' and ``0-class'' phrases can be found in the Appendix.

\section{Results and Analysis}
\label{sec:results_and_analysis}
Table \ref{tab:main-results} shows the results of running all 3 models on both the zero-shot and the few-shot settings. We make the following observations below:

\begin{enumerate}[wide, labelwidth=!, labelindent=0pt]
\item \textbf{The overall results are generally quite good.} The standard RoBERTa-large setting (i.e. no $k$-shot/zero-shot) attains an F1 score of 0.836, while a zero-shot model attains an F1 score of 0.740, which is a relatively high F1 score, considering that all the examples during test time were unseen during training. This shows that the model is able to learn something beyond simply just memorizing the PETs during training, and that it is able to somewhat capture the essence of what makes a phrase euphemistic. Perhaps it is able to track discrepancies in sentiment \cite{lee-etal-2022-searching} or discrepancies in other features such as politeness. At this point, it is difficult to discern exactly why the zero-shot performance is good, and it is an interesting point to explore further in the future.

\item \textbf{The ``bodily functions'' category performs quite poorly, while the ``substances'' category performs quite well.} For the ``bodily functions'', this can easily be explained by the dataset size and test set quality. Among the categories, ``bodily functions'' by far had the least number of test examples at 26 (see Table \ref{tab:dataset-statistics}). In fact, there appears to be some correlation between the performance and the size of the test set, as the ``sexual activity'' category (second-smallest test set) also exhibits relatively poor performance. In addition, the ``bodily functions'' category has a disproportionately high number of items with label 1 (i.e. euphemistic usage), which can skew the F1-score quite a bit. Observe that the macro precision is 0.5 for all 3 models, which tends to happen when the distribution is very skewed and gets a precision of exactly 1.0 for one class and exactly 0.0 for the other class. Meanwhile, for the ``substances'' category performing well, we speculate that this could be because a lot of these words are quite common. Words like ``weed'' and ``sober'' are used quite commonly, as opposed to other euphemisms, which are less commonly used in everyday conversations (e.g. ``ethnic cleansing'' is a rare phrase).

\item \textbf{GPT-3 generally performed quite poorly. Furthermore, GPT-3 performance seems to be independent of category.} For all 7 categories, as well as the randomly sampled zero-shot set, the GPT-3 model has F1 scores between 0.47 and 0.50 for almost all of them. This is a sharp contrast with the RoBERTa model, which varies quite significantly depending on the category. In addition, the GPT-3 performance is much lower than the RoBERTa performance. We hypothesize that this can be solved with additional prompt engineering or prompt tuning \cite{li-liang-2021-prefix, lester-etal-2021-power}. This poor performance can also be a possible cause for the lack of category dependence -- perhaps the model is not good enough to discern the subtle differences between these categories in the same way that the RoBERTa models do.

\item \textbf{The few-shot performance is better than the zero-shot performance. The 3-shot performance for RoBERTa is almost at level of training the standard model.} This should not come as a surprise, since having at least 1 appearance in the training set is already quite a lot of information provided to the model. Furthermore, the initial dataset had 1965 sentences split across 129 unique PETs, which averages out to around 15 sentences per PET. It is thus notable that being shown 3 examples gives almost the same performance as being shown 15 examples. This suggests that maybe a lot of the learning happens in the early stages, or that many sentences are actually redundant for training purposes. Another interesting area for future exploration would be in trying to find which sentences are the most ``instructive'' and hence best included within the training set for few-shot settings. 

\item \textbf{The GPT-3 model greatly benefited from the few-shot setting.} Comparing the $k=1$ and $k=3$ GPT-3 results with the zero-shot results, we see that there is a marked increase in performance when a few examples were given as prompts to GPT-3. This is consistent with the findings of \citet{gpt3} regarding GPT-3's capacity to perform in-context learning. This also makes intuitive sense, as simply providing GPT-3 with a single sentence to classify with no additional context can be quite difficult. In the first place, the GPT-3 model may not even fully know the task from a zero-shot setting. With just 3 examples, the F1 score increases from 0.507 to 0.617, which is a significant increase.

\end{enumerate}

\section{Conclusion and Future Work}
\label{sec:conclusion_future_work}
In this paper, we explored zero-shot and few-shot settings for the Euphemism Detection task. We formulated the problem settings and crafted zero-shot and few-shot datasets from the EMNLP 2022 FigLang Workshop Euphemism Shared Task dataset. We tried two type of models, namely RoBERTa and GPT-3. We found promising results that these language models (especially fine-tuned RoBERTa) were able to perform quite well, even on completely unseen euphemistic terms.

While our results were overall good, the results for GPT-3 were quite poor. In the future, we believe that further prompt engineering or prompt tuning will definitely be helpful in improving the performance of GPT-3 \cite{li-liang-2021-prefix, lester-etal-2021-power}. Furthermore, this idea of few-shot and zero-shot detection is not exclusive to euphemisms. We believe that checking the performance of language models to classify unseen examples is something that will be important to check for a lot of figures of speech and will be important in our quest to process and generate figurative text.

\section*{Limitations}
As mentioned in the body, a key limitation to our work is the lack of prompt engineering or prompt tuning. We tried some manually crafted prompts, but this does not seem to be enough to get GPT-3 to perform at the level it is expected to.

\bibliography{anthology,custom}
\bibliographystyle{acl_natbib}

\newpage
\appendix
\section{GPT3 Implementation: Positive and negative phrases}
\label{appendix:dataset-details}
Note that all sentences are converted to lowercase first before doing a search with these phrase lists. The ``1-class'' phrases and ``0-class'' phrases are shown below:

``1-class'': ["is used euphemistically", "can be used euphemistically", "is being used euphemistically", "may be used euphemistically", "might be used euphemistically", "is a euphemism", 
"is used as a euphemism", "is being used as a euphemism", "can be used as a euphemism", "may be used as a euphemism", "might be used as a euphemism",
"appears to be a euphemism", "appears to be used euphemistically", "could be used euphemistically", "could be used as a euphemism", "could be a euphemism",
"is considered a euphemism", "could be considered a euphemism", "can be considered a euphemism", "could be seen as a euphemism", "can be seen as a euphemism",
"could be considered euphemistic", "can be considered euphemistic",
"i think so", "i believe so"]

``0-class'': ["not used euphemistically", "not being used euphemistically", "not a euphemism", "not used as a euphemism", "not being used as a euphemism",
"does not appear to be a euphemism", "does not appear to be used euphemistically",
"i don't think so", "i don't believe so", "i do not think so"]

\end{document}